\documentclass[letterpaper, 10 pt, conference]{ieeeconf}  

\IEEEoverridecommandlockouts                              

\usepackage{graphics} 
\usepackage{epsfig} 
\usepackage{mathptmx} 
\usepackage{times} 
\usepackage{amsmath} 
\usepackage{amssymb}  
\usepackage{algorithm} 
\usepackage{cite}
\usepackage{algpseudocode} 
\usepackage{lipsum}
\usepackage{graphicx}
\usepackage[dvipsnames]{xcolor}
\usepackage{eucal}
\usepackage{mathtools,leftidx}
\usepackage{subcaption}
\usepackage{wrapfig}

\usepackage{multirow}

\usepackage{color, colortbl}
\usepackage[bookmarks=true]{hyperref}

\usepackage[utf8]{inputenc}
\usepackage{fontenc}
\usepackage{siunitx}
\usepackage{tikz}
\usepackage{textcomp}
\newcommand\footnoteref[1]{\protected@xdef\@thefnmark{\ref{#1}}\@footnotemark}

\makeatletter
\newcommand*\titleheader[1]{\gdef\@titleheader{#1}}
\AtBeginDocument{%
  \let\st@red@title\@title
  \def\@title{%
    \bgroup\normalfont\large\centering\@titleheader\par\egroup
    \vskip1.5em\st@red@title}
}
\makeatother

\title{\LARGE \bf
GP-Frontier for Local Mapless Navigation
}

\author{Mahmoud Ali and Lantao Liu
\thanks{$^{1}$Mahmoud Ali and Lantao Liu are with the Luddy School of Informatics, Computing, and Engineering, Indiana University, Bloomington, IN 47408 USA, {\tt\small \{alimaa, lantao\}@iu.edu}}
}

\titleheader{ \textcolor{gray}{This paper has been accepted for publication at 2023 IEEE International Conference on Robotics and Automation} \textcolor{blue}{\href{https://www.icra2023.org/}{(ICRA 2023)}} }

\begin{document}

\maketitle
\thispagestyle{empty}
\pagestyle{empty}

\begin{abstract}

We propose a new frontier concept called the Gaussian Process Frontier (GP-Frontier) that can be used to locally navigate a robot towards a goal without building a map. The GP-Frontier is built on the uncertainty assessment of an efficient variant of sparse Gaussian Process.  
Based only on local ranging sensing measurement, the GP-Frontier can be used for navigation in both known and unknown environments. The proposed method is validated through intensive evaluations, and the results show that the GP-Frontier can navigate the robot in a safe and persistent way, i.e., the robot moves in the most open space (thus reducing the risk of collision) without relying on a map or a path planner. 
A supplementary video that demonstrates the robot navigation behavior is available at \url{https://youtu.be/ndpqTNYqGfw}. 
\end{abstract}

\vspace{-5pt}
\section{Introduction}
The concept of \textit{frontier-based} navigation 
was firstly proposed by \cite{yamauchi1997frontier}. The frontiers are defined as the boundary grids between unexplored
and explored space, and they usually appear on the maximum range of sensing (or the ``edge of the sensing sweep" that does not return any obstacle detection). 
Since the frontiers are on the boundary between known free space and unknown space that has not yet been sensed, the frontiers are used to navigate the robot to further scan the unknown space, and repeatedly, the known (mapped) territory will continuously expand by pushing the boundary toward the unknown areas, leading to interesting exploration behavior.  
When there is no new frontier left, the unknown space exploration is deemed complete. 


There are some drawbacks for existing frontier-based navigation.  
The first important issue for the existing frontier concept lies in the inappropriate assumption that the frontiers (discrete boundary grids) are independent to each other. 
In the real world, space has continuity and correlation, and this property has been ignored. 
The second issue is the reliance between frontiers and a map. The conventional frontiers are defined, and thus dependent, on a map (data) structure and, usually, the mapping process. Consequently, existing work typically leverages frontiers for unknown space exploration and map construction where the spatial coverage is an important goal.  In many practical tasks, the robot does not need to explore or map the space, but simply needs to continuously navigate in the environment and might repetitively revisit the same locations that have been visited  many times before (e.g., patrolling,  surveillance).

To tackle the above two issues simultaneously, 
we propose a new frontier concept called GP-Frontier based on 
a novel compact form of perception model 
constructed with an onboard ranging sensor. 
The GP-Frontier can guide and navigate the robot in known or unknown space, with or without a goal. 
Different from existing work, our solution does not rely on any map and can navigate the robot continuously and safely.  
This is achieved by 
utilizing the variational sparse  Gaussian Process (VSGP) to build a local occupancy surface, where all the 3D occupied points observed by the ranging sensor are projected onto a 2D circular surface 
that considers the correlations of the observed points and the uncertainty of the regression model.
Frontiers selected thus are typically located in the most open space, which is important for safe local navigation.  
In other words, the GP-Frontier shows unique navigation capabilities because its foundation is based on spatial correlation and uncertainty assessment, which is very different from conventional frontier definitions.  
\begin{figure} 
    \centering
  \subfloat[\label{fig_subgoal_a}]{
      \includegraphics[scale=0.14]{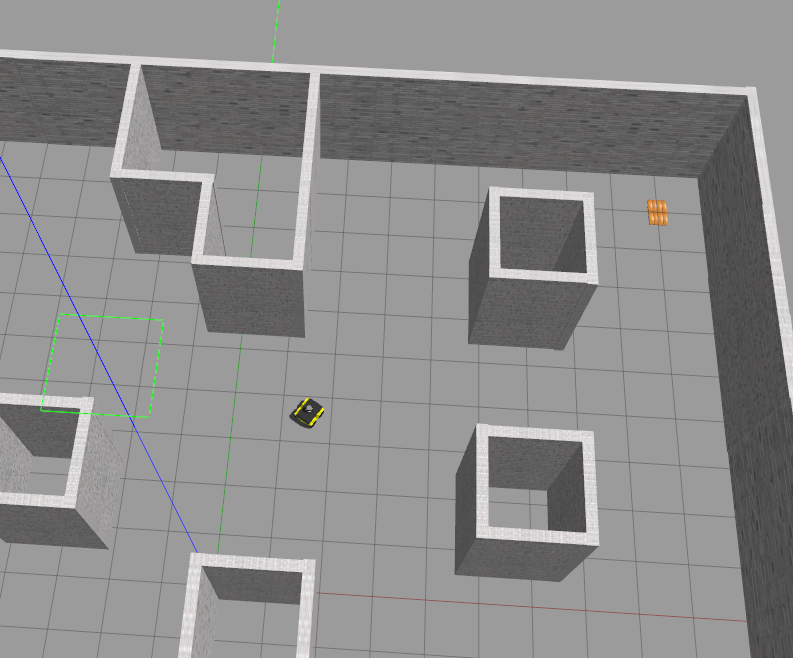} } \
   \subfloat[\label{fig_subgoal_d}]{
      \includegraphics[scale=0.235]{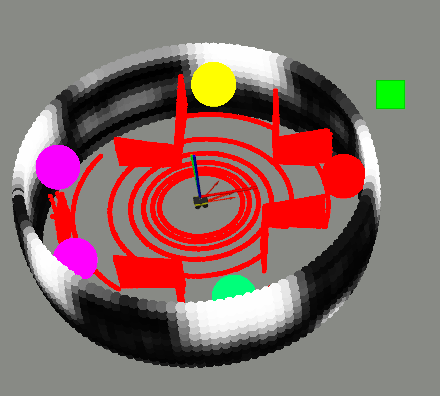} }  
   \includegraphics[width=0.04\linewidth,height=1.3in]{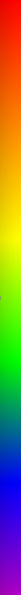}  
      \vspace{-10pt}
  \caption{\small GP-Frontier: (a) a simulation scene where a Jackal robot is surrounded by sparse obstacles; (b) the colored small spheres represent the GP-Frontiers (equivalent to local gaps or sub-goals) detected by our method; the colors reflect the cost (assigned by a cost function) for each sub-goal based on the final navigation goal (shown as a green squared). Raw pointcloud is shown in red in (b).\vspace{-15pt}}
  \label{fig_subgoal} 
\end{figure}
Specifically, in this paper we present the GP-Frontier and its local navigation method by using only an onboard ranging sensor -- we use LiDAR as an example. 
The local observation is represented as an occupancy surface, where all the 3D occupied points observed by the LiDAR are projected onto a 2D circular surface modeled with VSGP. 
Then the uncertainty of the VSGP model is used to detect all potential GP-Frontiers (sub-goals) around the robot. Based on the distance and direction of each GP-Frontier relative to both the robot and the final goal, a cost function selects the most promising GP-Frontier that will drive the robot to the final goal. At the last step, a motion command is generated as a function of both the distance and direction of the selected GP-Frontier relative to the robot, see Fig.~\ref{fig_subgoal} for an illustration.
\vspace{-10pt}
\section{Related Work} \label{related_work}
\vspace{-3pt}
The frontier based navigation and exploration has been studied~\cite{yamauchi1997frontier, yamauchi1998frontier, holz2010evaluating}.
Most existing frontier exploration is coupled with a mapping method. For example, A 3D frontier detection method, named Stochastic Differential Equation-based Exploration algorithm~(SDEE) was proposed in \cite{shen2012stochastic} 
where the authors demonstrate the efficiency on quadrotors in a 3D map representation. 
The frontier-based exploration has also been extended to UAV-UGV collaborations\cite{butzke20153,wang2018collaborative}, as well as multi-agent cooperative exploration \cite{burgard2000collaborative}  which integrates both the cost of reaching a target and also the utility of target points. 
Recently, exploration at high speeds leveraged the frontiers where a goal frontier is locally selected from the robot's field of view \cite{cieslewski2017rapid} so that the change in velocity to reach the local goal is minimized.
Another framework in the same line
is the gap-based method, where a gap is a free space between two obstacles that the robot can pass by. The first method, Nearness Diagram, was introduced by~\cite{1266644}, then many variants were developed based on this approach. As stated in~\cite{mujahed2018admissible}, ND-based methods showed undesired oscillatory motion. To solve the gap-based method's limitations, researchers in robotics developed schemes based on the geometry of the gap. 
The Follow-the-Gap Method (FGM)~\cite{sezer2012novel}, selects one of the detected gaps based on the gap area and calculates the robot heading based on the direction of the gap center relative to both the robot, and the final goal. Sub-goal seeking approach~\cite{ye2009sub} defines a cost for each sub-goal as a function of both the sub-goal and goal heading errors with respect to the robot heading, then it selects the sub-goal with the minimum cost (error). 
Our work is most related to 
the Admissible Gap (AG) approach as both aim to address  reactive collision avoidance~\cite{mujahed2018admissible}. AG, an iterative algorithm, considers the exact shape and kinematic constraints of the robot, finds the possible admissible gaps, and then chooses the nearest gap as the goal, thus decreasing motion oscillations, path length, and safety risks.

Distinct from AG, our work is based on a variant of GP. The standard form of GP is known to suffer from few limitations~\cite{rasmussen2002infinite}, the most significant one is the cubic computational complexity of a vanilla implementation. 
However, different methods -collectively known as Sparse Gaussian Process (SGP)- tackle the computation complexity of GP~\cite{snelson2006sparse,titsias2009variational,hoang2015unifying,bauer2016understanding}. For instance, online SGP~\cite{ma2017informative} has been proposed to reduce the computational demand associated with modeling large data sets using GP. Also, a Mixture of GPs was adopted by \cite{kim2012building, luo2018adaptive} to capture non-stationarity environmental attributes.
Naturally, many GP-based occupancy mapping methods were intensively explored in the literature \cite{GPOM, jadidi2018gaussian, kim2012building}. 
In this work, we choose the VSGP due to its efficiency in computation and sensing representation which is important for real-time robot navigation. 

\section{Methodology} \label{methodology}

\subsection{{Occupancy Surface Construction with Sensor Data}}
We define the LiDAR local observation (pointcloud) in the spherical coordinate system, where any point is represented by the tuple $(\theta, \alpha, r)$ which describes the azimuth, elevation, and distance (radius) values of the 3D point with respect to the sensor origin, respectively. 
The occupied points observed by LiDAR are projected onto the  occupancy surface, which is a circular surface around the sensor origin with a predefined radius $r_{oc}$~\cite{realtime-expo}.
Any point on the occupancy surface is defined by two attributes $\theta$ and $\alpha$, where $\theta$'s values range from $-\pi$ to $\pi$ and $\alpha$'s values depend on the LiDAR's  vertical Field of View (FoV)  (for VLP16 $\alpha$ range is  $-15^{\circ}$ to $15^{\circ}$). Each point  $\mathbf{x}_i = (\theta_i,  \alpha_i)$ on the occupancy surface is assigned an occupancy value $oc_i = r_{oc} - r_i$, where $r_i$ is the point radius. All points that have a radius $r$ shorter than the occupancy surface radius $r_{oc}$ ($r < r_{oc}$) form the {\em occupied space} of the occupancy surface; the rest of the points on the surface with radius $r$ greater than or equal to the occupancy radius $(r >= r_{oc})$ are considered as the {\em free space} of the occupancy surface ($oc=0$), see Fig.~\ref{fig_vsgp_mdl_a}.
\vspace{-2pt}
\subsection{{Occupancy Surface Representation with VSGP}} \label{vsgp_oc_mdl}
\begin{figure} 
\vspace{5pt}
    \centering
   \subfloat[\label{fig_vsgp_mdl_a}]{
      \includegraphics[scale=0.14]{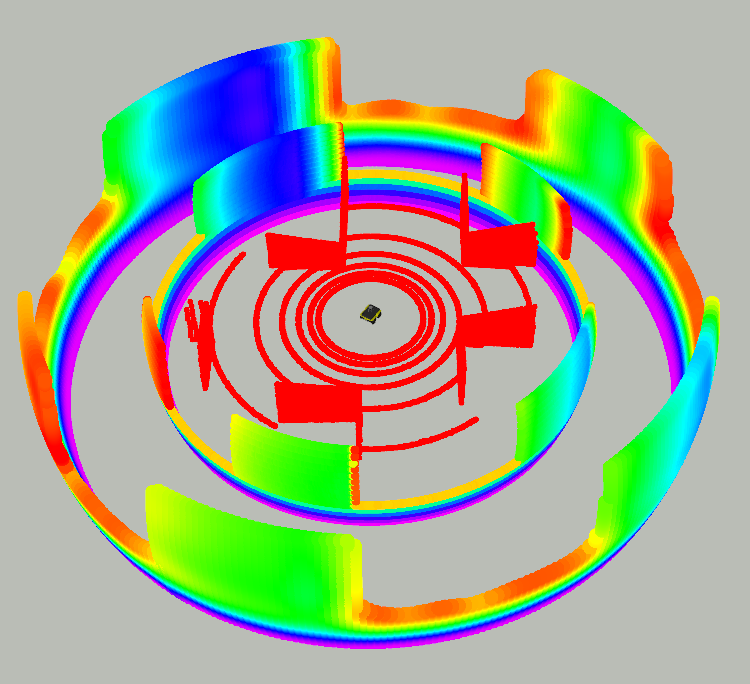} }    
  \subfloat[\label{fig_vsgp_mdl_b}]{
      \includegraphics[scale=0.163]{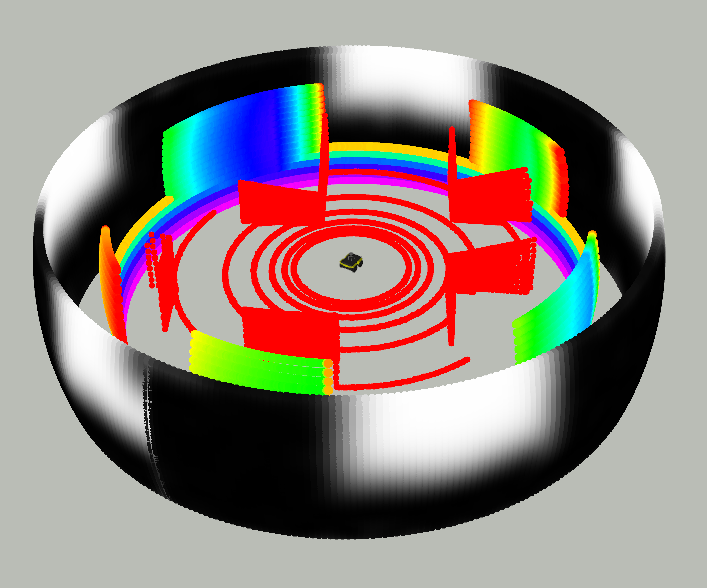} }
         \includegraphics[width=0.04\linewidth,height=1.33in]{occ_srfc/bar.png}  
      \vspace{-8pt}
  \caption{\small Occupancy Surface: raw pointcloud of the scene in Fig.~\ref{fig_subgoal_a} is shown in red in both (a) and (b); the inner circular surface in (a) and (b) represents the original occupancy surface; while the outer circular surface in (a) represents the reconstructed occupancy surface predicted by our VSGP occupancy model;  the outer circular surface in (b) represents the uncertainty (variance) of the reconstructed occupancy surface. Both original and reconstructed occupancy surfaces have the same radius, but for better visualization, we depict the reconstructed surface with an extended radius.  \vspace{-15pt}}
  \label{fig_vsgp_mdl} 
\end{figure}
The occupied points on the occupancy surface are transformed to a training data set $\mathcal D = \left\{\left(\mathbf{x}_{i}, y_{i}\right)\right\}_{i=1}^{n}$ with an $n$ input points, $\mathbf{x}_i=(\theta_i, \alpha_i)$, and their corresponding occupancy values, $y_i=oc_i$. 
$\mathcal D$ is then used to train a 2D VSGP occupancy model $f(\mathbf{x}_i)$ which describes the probability of occupancy over the occupancy surface as follow:
\vspace{-5pt}
\begin{equation}
    \begin{gathered}
    f(\mathbf{x}) \sim {VSGP}\left(m(\mathbf{x}), k_{RQ}\left(\mathbf{x}, \mathbf{x}^{\prime}\right)\right),  \\
    k_{\mathrm{RQ}}\left(\mathbf{x}, \mathbf{x}^{\prime}\right)=\sigma^{2}\left(1+\frac{\left(\mathbf{x}-\mathbf{x}^{\prime}\right)^{2}}{2 \alpha \ell^{2}}\right)^{-\alpha},
    \end{gathered}
    \label{eq_mean_kernel_vsgp}    
\end{equation}
where $k_{RQ}\left(\mathbf{x}, \mathbf{x}^{\prime}\right)$ is the Rational Quadratics (RQ) co-variance function (known as a kernel) with signal variance $\sigma^{2}$, length-scale $l$, and  relative weighting factor $\alpha$. 
A Gaussian noise  $\epsilon$ is added to the model to reflect the measurement's noise. Therefore, the probability of occupancy $y_i$ of any point $\mathbf{x}_i$ is defined as $y_i=f(\mathbf{x}_i)+\epsilon$, where $\epsilon$ is sampled from a Gaussian distribution $\mathcal{N}\left(0, \sigma_{n}^{2}\right)$ with a variance $\sigma_{n}^{2}$.
The {GP} posterior is represented by the posterior mean $m_{\mathbf{y}}(\mathbf{x})$ and posterior covariance $ k_{\mathbf{y}}\left(\mathbf{x}, \mathbf{x}^{\prime}\right)$~\cite{titsias2009variational} as
\vspace{-5pt}
\begin{equation}
    \begin{gathered}
    m_{\mathbf{y}}(\mathbf{x})=K_{\mathbf{x} n}\left(\sigma^{2} I+K_{n n}\right)^{-1} \mathbf{y}, \\
    k_{\mathbf{y}}\left(\mathbf{x}, \mathbf{x}^{\prime}\right)=k\left(\mathbf{x}, \mathbf{x}^{\prime}\right)-K_{\mathbf{x} n}\left(\sigma^{2} I+K_{n n}\right)^{-1} K_{n \mathbf{x}^{\prime}},
    \end{gathered}
    \label{eq_posterior_mean_kernel_full_gp}
\end{equation}
where $\mathbf{y}= \left\{y_{i}\right\}_{i=1}^{n}$, 
$K_{nn}$ is $n \times n$ co-variance matrix of the inputs, $K_{xn}$ is n-dimensional row vector of kernel function values between $\mathbf{x}$ and the inputs, and $K_{nx} = K_{xn}^T$. 

For accurate GP prediction, the kernel parameters $\Theta$ and the noise variance $\sigma_n^2$ should be correctly tuned. 
In this work, we use a sparse variant of GP where only $m$ input points ({\em inducing points} $X_m$) are considered to describe the entire training data. 
The values of the latent function $f(\mathbf{x})$ at $X_m$ are called the {inducing variables} $f_m$.
Specifically, we leverage the VSGP framework proposed in~\cite{titsias2009variational} that jointly estimates the kernel hyperparameters $\Theta$ and selects the inducing inputs $X_m$ by approximating the GP posterior $p(f | y, \theta)$ with a variational posterior distribution. 
The computation complexity of SGP, $\mathcal{O}(nm^2)$ is dependent on the number of inducing points $m$, however, it is always less than the full GP's computation complexity, $\mathcal{O}(n^3)$.  
More details of the VSGP can be found in  Titsias's seminal work~\cite{titsias2009variational}. 

\ A zero-mean function is used to represent a zero-occupancy prior over the surface, which means that before acquiring any observation, the occupancy of all points is set to zero. 
The RQ kernel is selected to form a flexible GP prior with a set of functions that vary across different length scales. The resolution of the LiDAR along the azimuth and elevation axes is used to initiate different length scales along both axes.
The VSGP optimizes both the variational parameters (inducing inputs $X_m$) and the hyperparameters $\Theta$ through a variational Expectation-Maximization (EM) algorithm. In our model, the limited input domain of the VSGP ( $-\pi<\theta<\pi$ and $\alpha_{min}<\alpha <\alpha_{max}$) is exploited to initialize the inducing inputs $X_m$ at evenly distributed points on the {\em occupied part} of the occupancy surface. At each E-step, a different set of points is chosen to maximize the variational objective function 
while the hyperparameters are updated during the M-step~\cite{titsias2009variational}.


\ After estimating the hyperparameters and the inducing points, the VGSP occupancy model is used to predict the probability of occupancy $y^*$ for any point $\mathbf{x}^*$ on the occupancy surface by employing the GP prediction equation
\begin{equation}
    p(y_* | \textbf{y}) = \mathcal{N}(y_* | m_{\textbf{y}}(\textbf{x}_*), k_\textbf{y}(\textbf{x}_*,\textbf{x}_*) + \sigma^2_n). 
    \label{eq_predictive_eq_full_gp}
\end{equation} 
Fig.~\ref{fig_vsgp_mdl_a} shows the \textit{original occupancy surface} (inner circular surface) and the \textit{reconstructed occupancy} surface predicted by the VSGP occupancy model (the outer circular surface). Both the original and the predicted occupancy surfaces in Fig.~\ref{fig_vsgp_mdl_a} are color-coded, where warm colors reflect low occupancy; low occupancy means higher radius values as $r_i = r_{oc} - oc_i$. 
Therefore, for any direction defined by the azimuth $\theta$ and elevation $\alpha$ angels, our VSGP occupancy model estimates the distance $r_i$ to the obstacle along that direction. 
Our previous study~\cite{ali2023light} investigated the accuracy of the VSGP model, where the results demonstrated that a VSGP with 400 inducing points results in an average error of around \SI{12}{\centi\metre} in the reconstructed point cloud.

\subsection{{GP-Frontiers for Local Navigation} }
A key benefit of using the GP and its variations over other modeling methods is their ability to quantify the uncertainty, or variance, associated with the predicted value at any given query point. 
For each point on the reconstructed occupancy surface, the occupancy $\mu_{oc}$ predicted by the VSGP model is associated with a variance value $\sigma_{oc}$.
While the occupancy surface can be considered as a \textit{local} 3D map -projected on a 2D circular surface- of the robot locality (see Fig. \ref{fig_vsgp_mdl_a}), the variance $\sigma_{oc}$ associated with the reconstructed occupancy surface can be considered as a local uncertainty-map of the robot locality, see Fig.~\ref{fig_vsgp_mdl_b}.
The \textit{variance surface} (grey-coded surface in Fig.~\ref{fig_vsgp_mdl_b}) generated by the VSGP defines the certain and uncertain regions in the robot locality. Therefore, any region on the variance surface can be classified as known space (low uncertainty regions) shown as black regions on the variance surface, or unknown space (high uncertain regions) which we call a GP-Frontier candidate shown as white regions on the variance surface, see Fig.~\ref{fig_vsgp_mdl_b}.

The well-known frontier concept introduced in \cite{yamauchi1997frontier} is associated with exploration and occupancy map building, however, in this paper, we do not consider any kind of global map or map building techniques.
Instead, our proposed GP-Frontier exploits the correlations of the observed points and is selected by leveraging the uncertainty of the VSGP occupancy model. GP-Frontiers on the occupancy surface are typically located in either the most open space (non-occupied space) or the region in the occupied space that has a large discontinuity on the occupancy value. This large discontinuity is explained as the gaps between different obstacles in the occupied space, which are also considered as GP-Frontier candidates.
In this context, we only consider one full sensor scan/observation 
to represent the occupancy of the robot locality as a VSGP occupancy model and define local navigation points (i.e. GP-Frontiers) based on the VSGP uncertainty. Any region on the variance surface with a variance that is higher than a threshold $V_{th}$ is considered a GP-Frontier. 
Actually, The variance associated with the occupancy surface varies for each observation and is influenced by the quantity and the arrangement of the observed points on the surface.
Consequently, the variance threshold $V_{th}$ is determined as a variable that varies with the variance distribution across the surface, $V_{th}= K_m*v_m$ where $v_m$ is the variance distribution mean and $K_m$ is a tuning parameter.

Formally, a GP-Frontier $f_i=(\theta_{f_i}, \alpha_{f_i}, r_{f_i})$ is defined by its centroid point on the variance surface $\mathbf{x}_{f_i}= (\theta_{f_i}, \alpha_{f_i})$, and the distance $r_{f_i}$ between the GP-Frontier centroid and the occupancy surface origin. $r_{f_i}$ is estimated using the VSGP occupancy model, 
where $oc_{f_i}=vgsp((\theta_{f_i}, \alpha_{f_i}))$ and $r_{f_i}=r_{oc} - oc_{f_i}$. In practice, for 2D navigation, GP-Frontier can be defined as $f_i=(\theta_{f_i}, 0, r_{f_i})$ because $\alpha_{f_i}$ is a constant ($\alpha_{f_i} =\alpha_{xy} = 0$), where $\alpha_{xy}$ is the elevation angel of the 2D XY-plane. $\theta_{f_i}$ is used to define the GP-Frontier direction with respect to the robot heading, considering the transformation between the robot frame $\mathcal{R}$ and the LiDAR fame is known. 
The cartesian coordinates of GP-Frontier $f_i$ in a global world frame $\mathcal{W}$ are calculated as $(x^{w}_{f_i}, y^{w}_{f_i}) = \prescript{W}{}{\textbf{T}}_{R} (x^{R}_{f_i}, y^{R}_{f_i})$ where $\prescript{W}{}{\textbf{T}}_{R}$ is the transformation between $\mathcal{R}$ and $\mathcal{W}$ (robot localization) and  $(x^{R}_{f_i}, y^{R}_{f_i})$ are the cartesian coordinates of the GP-Frontier $f_i$ in $\mathcal{R}$ which correspond to the GP-Frontier spherical coordinates $(\theta_{f_i}, 0, r_{f_i})$.
\subsection{{GP-Frontier for Goal-Oriented Navigation} }
To navigate towards a given final goal $\mathbf{g} = (x_g, y_g)$ in $\mathcal{W}$, local navigation approaches use different criteria to select one sub-goal (i.e., an ideal GP-Frontier $f^*$, or a gap) from the local GP-Frontiers.  
Our approach combines both distance and direction criteria. A cost function $C$ is used to select only one GP-Frontier $f^*$ to act as the next navigation sub-goal. The proposed cost function $C$ adopts the distance criteria proposed in~\cite{yan2020mapless} and the direction criteria proposed in~\cite{ye2009sub}: 
\vspace{-10pt}
\begin{equation}
 \label{eq_cost_fun} 
    \begin{aligned}
C\left(f_{i}\right) &=  k_{dst}  d_{sum} + k_{dir} \theta_{f_i}^2 ,  \\
d_{sum} &= r_{f_i} + \sqrt{(x_g-x^w_{f_i})^2+(y_g-y^w_{f_i})^2},\\
f^{*} &=\operatorname{arg} \min _{f_{i} \in \mathcal{F}}\left(C\left(f_{i}\right)\right), 
\end{aligned}     
\end{equation}
where $k_{dst}$, $k_{dir}$ are weighting factors.
Integrating both distance and direction criteria decreases the chance of getting stuck in local minima. More discussion is provided in Sec.~\ref{sec_sim_results}.

 

Finally, a motion command $(v, \omega)$ is generated to drive the robot towards the center of the selected GP-Frontier $f^*$. 
The linear velocity $v$ varies directly with the distance to the sub-goal ($r_{f^*}$) and inversely with the direction to the sub-goal $\theta_{f^*}$;  $v = k_a r_{f^*} - k_b \|\theta_{f^*}\|$. The angular velocity $\omega$ is proportional to the direction of the sub-goal; $\omega = k_c \theta_{f^*} $. $k_a, k_b$ and $k_c$ are tunable coefficients.
If the final goal is inside the local FoV of the robot, then the motion command drives the robot directly towards the goal, otherwise, the motion command drives the robot towards the selected sub-goal $f^*$.


\begin{algorithm}
{\small
	\caption{
 GP-Frontier Local Mapless Navigation \\
        \textbf{INPUT}  : LiDAR Observation (Pointcloud (PCL) ) \\
        \textbf{OUTPUT}: Motion Command
	}  
	\begin{algorithmic}[1]
	    \State $X_m$: Inducing Points
	    \State $\mathbf{X^*} \gets (\theta, \alpha)$: 2D Variance Grid 
            \State $\mathbf{g} = (x_g, y_g)$: Navigation goal 
            \While { New Sensor Observation (PCL) }
	        \State $(\theta_i, \alpha_i, r_i) \gets$ \textit{Catersian2Spherical}(PCL)
	        \State $oc_i=r_{oc}-r_i$
	        \State $\mathbf{x}_i= (\theta_i, \alpha_i) $,  $y_i=oc_i$
                \State $\mathcal D = \left\{\left(\mathbf{x}_{i}, y_{i}\right)\right\}_{i=1}^{n} \gets \mathbf{x}_i, y_i$
                \State  $f(\mathbf{x}) \sim {VSGP}\left(m(\mathbf{x}), k_{RQ}\left(\mathbf{x}, \mathbf{x}^{\prime}\right)\right)$        
		    \State \textit{Optimize}( $\Theta, X_m$) $\gets \mathcal{D}$
                \State $\mu_{oc}, \sigma_{oc} \gets \mathcal{N}(\textbf{y}^* | m_{\textbf{y}}(\textbf{X}^*), k_\textbf{y}(\textbf{X}^*,\textbf{X}^*) + \sigma^2_n)$
		    \State $v_m\gets$\textit{Mean}($\sigma_{oc}$) 
		    \State $V_{th} \gets K_m v_m$ 
		    \State GP-Frontiers $\gets$  $\sigma_{oc}>V_{th}$
			\For { each GPF $fi$ in GPFs}
                \State $\mathbf{x}_{f_i} = (\theta_{f_i}, \alpha_{f_i}) \gets$ \textit{CentroidOfGP}-Frontier    
                \State ${oc}_{f_i}, \sigma_{f_i} \gets \mathcal{N}(\textbf{y}^* | m_{\textbf{y}}(\textbf{x}_{f_i}), k_\textbf{y}(\textbf{x}_{f_i},\textbf{x}_{f_i}) + \sigma^2_n)$
                
                \State $r_{f_i} \gets r_{oc} - oc_{f_i}$   
                \State $(x^R_{f_i}, y^R_{f_i}, 0)$ $\gets$ \textit{Spherical2Cartesian}($\theta_{f_i}, 0, r_{f_i}$)
                \State $(x^w_{f_i}, y^w_{f_i}, 0)$ $\gets$ $\prescript{W}{}{\textbf{T}}_{R} (x^R_{f_i}, y^R_{f_i}, 0)$
                \State $d_{sum} = r_{f_i} + \sqrt{(x_g-x^w_{f_i})^2+(y_g-y^w_{f_i})^2}$
                \State $C\left(f_{i}\right) =  k_{dst}  d_{sum} + k_{dir} \theta_{f_i}^2$
			\EndFor
			\State $f^{*} \gets \operatorname{arg} \min _{f_{i} \in \mathcal{F}}\left(C\left(f_{i}\right)\right),$
			\State  $v = k_a r_{f^*} - k_b \|\theta_{f^*}\|$
			\State $\omega  \gets k_c \theta_{f^*} $
		\EndWhile
		\end{algorithmic} 
	\label{pseudo_algorithm}
}
\end{algorithm}
\begin{figure}       \vspace{5pt}
    \centering
    \subfloat[Environment \textbf{A}\label{fig_exp_md}]{%
      \includegraphics[width=0.45\linewidth,height=1.4in]{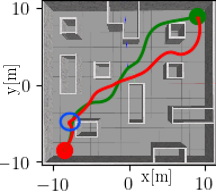}} \hfill
     \subfloat[Environment \textbf{B}\label{fig_exp_x}]{%
      \includegraphics[width=0.45\linewidth,height=1.4in]{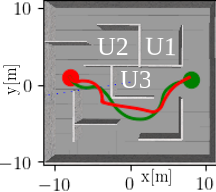} } \hfill
      \vspace{-5pt}
  \caption{\small Simulation experiments. (a) MD; (a) X; 
  where green paths generated by GP-Frontier and red paths generated by AG.
  \vspace{-10pt}}
  \label{fig_env} 
\end{figure}
\begin{figure} 
    \centering
   \subfloat[Runtime\label{fig_exp_md_time}]{
      \includegraphics[scale=0.57]{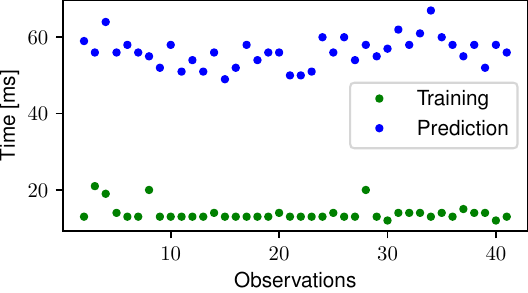} }
  \subfloat[Angular velocity $\omega$\label{fig_exp_md_w}]{ \includegraphics[scale=0.43]{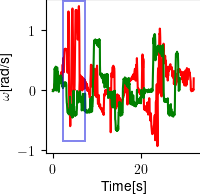} } 
  \vspace{-5pt}
  \caption{\small Experiment MD. 
  \vspace{-10pt}} 
  \label{fig_exp_timimg} 
\end{figure}
\begin{figure*} \vspace{5pt}
    \subfloat[SU\label{fig_exp_env2_b}]{%
      \includegraphics[scale=0.71]{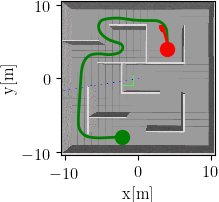} } \  
     \subfloat[CU\label{fig_exp_env2_c}]{%
      \includegraphics[scale=0.75]{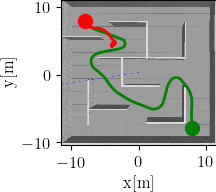}}  \ 
    \subfloat[GU\label{fig_exp_env2_d}]{%
      \includegraphics[scale=0.75]{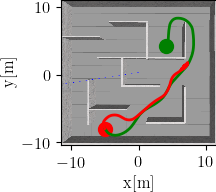}}  \ 
    \subfloat[$\omega$ corresponding to SU\label{fig_exp_env2_e}]{%
      \includegraphics[scale=0.73]{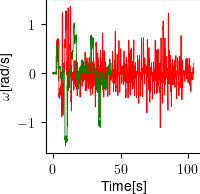}}  
  \vspace{-5pt}
  \caption{\small 
  Simulation experiments SU, CU, and GU. Green paths generated by GP-Frontier, while red paths generated by AG. \vspace{-15pt} 
  }
  \label{fig_exp_env2}
\end{figure*}
\vspace{10pt}
\section{Experimental Design and Results}
\subsection{Simulation Setup} \label{simulation_experiment}
The proposed GP-Frontier is built on top of GPFlow \cite{matthews2017gpflow} and executed in real-time. 
Real-time simulation in Gazebo and real-time demonstration were considered 
to evaluate the performance of our approach and to compare it to a baseline -- the AG method~\cite{mujahed2018admissible} that was published recently to address the same problem. 
During all of the simulation experiments, the maximum linear and angular velocities were set to \num{1.0} \si{\metre/\second} and \num{1.5} \si{\radian/\second} respectively.
LiDAR configurations were set to a $5 m$ maximum range (to have limited FoV compared to the environment size), a $5 Hz$ frequency, and a resolution of $(0.35^{\circ}, 2^\circ)$ along the azimuth and the elevation axis, respectively. 
The surface for occupancy was created with a radius $r_{oc}$ of \SI{5}{\metre} and a complete azimuth range from $-180^o$ to $180^o$, with an elevation height spanning from $0^o$ to $15^o$. Throughout the experiments, specific tuning parameters and constants were 
assigned as follows:  inducing points $X_m = 400$, variance threshold constant $K_m=0.4$, where the distance and direction  weighting factors $K_{dst}$ and $K_{dir}$ were set to \num{5} and \num{4}, respectively.
To predict occupancy, a 2D grid $\mathbf{X}^*$ is used to represent the surface, with a resolution identical to that of the LiDAR resolution along both the azimuth and elevation angles. 

Two environments \textbf{A} and \textbf{B}, each of them has an area of 20x20 meters, are used to evaluate the local navigation performance. Specifically, \textbf{A} is a cluttered environment with random obstacles, while \textbf{B} is a maze-like environment with 3 u-shaped rooms U1, U2, and U3, see Fig. \ref{fig_env}. 
Different experiments were designed to thoroughly evaluate the performance of the GP-Frontier and the AG methods. 
Specifically, 
i) MD: where the robot has to go along the main diagonal (MD) of environment \textbf{A}, see Fig.~\ref{fig_exp_md}.
ii) X: where the robot has to go parallel to the X-axis of environment \textbf{B}, see Fig.~\ref{fig_exp_x}.
iii) SU: where the starting pose is located inside U1, see Fig.~\ref{fig_exp_env2_b}.
iv) CU: where the robot has to cross U2 to reach the goal, see Fig.~\ref{fig_exp_env2_c}.
v) GU: where the goal is located inside U1, see Fig.~\ref{fig_exp_env2_d}. 
The starting pose and the final goal for each experiment are shown in Table~\ref{tab_performance}.
Five metrics are considered to evaluate the local navigation behavior~\cite{mujahed2018admissible}: \\
i) $T_{\text {tot }}$: total time to reach to the final goal $\textbf{g}$.\\
ii) $D_{\text {acc}}$: traveled distance to reach the final goal .\\
iii) $\mathrm{C}_{\text {chg }}$: trajectory curvature change measure that reflects the oscillations along the path.
\vspace{-3pt}
$$
\mathrm{C}_{\text {chg }}=\frac{1}{\mathrm{~T}_{\text {tot }}} \int_0^{\mathrm{T}_{\text {tot }}}\left|k^{\prime}(t)\right| d t, \quad k(t)=\left|\frac{\omega(t)}{v(t)}\right|.
$$
\\\vspace{-4pt}
iv) $\mathrm{J}_{\mathrm{acc}}$: accumulated jerk measure for  trajectory smoothness. \vspace{-4pt}
$$
\mathrm{J}_{\mathrm{acc}}=\frac{1}{\mathrm{~T}_{\text {tot }}} \int_0^{\mathrm{T}_{\text {tot }}}[\ddot{v}(t)]^2 d t. 
$$
\\\vspace{-4pt}
v) $\mathrm{R}_{\mathrm{obs}}$: risk measure that reflects  proximity to obstacles.
$$
\mathrm{R}_{\mathrm{obs}}=\int_0^{\mathrm{T}_{\mathrm{tot}}} \frac{1}{r_{\min }(t)} d t, 
$$
where $r_{\min }$ is the distance to the closest obstacle. For all these five metrics described above, a lower value indicates a better performance. 

\begin{table*}[t]
\vspace{5pt}
  \centering
  \begin{tabular}{cccccccccc} 
  \hline \\[-2ex]
 \textbf{Env.} & \textbf{Exp.} & \textbf{Starting pose} & \textbf{Goal} &  \textbf{Method} & $\mathbf{T_{tot}}$[\si{\second}] &  $\mathbf{D_{acc}}$ [\si{\metre}]& $\mathrm{\mathbf{J}}_{\mathrm{\textbf{acc}}}$[\si{\metre\per\cubic\second}]  & $\mathrm{\textbf{C}}_{\text {\textbf{chg} }}$[\si{\radian/\metre}]& $\mathrm{\textbf{R}}_{\mathrm{\textbf{obs}}}$[\si{\per\metre}]\\ \hline \\[-2ex]
  
\multirow{2}{*}{\textbf{A}} &  \multirow{2}{*}{MD}  &  \multirow{2}{*}{$(-8.5, -8.5, 45^o)$} & \multirow{2}{*}{$(8.5, 8.5, 45^o)$} &  GP-Frontier  &\textbf{ 28.68$\pm$1.2}  &  \textbf{28.2$\pm$0.74}  & \textbf{30.2$\pm$10.2} &  \textbf{18.14$\pm$18.9}  &  \textbf{19.76$\pm$1.1}\\ \\[-2ex]
& &   &  &  AG   & 34.11$\pm$4.3  &  28.4$\pm$0.66  & 74.1$\pm$9.7  &  76.19$\pm$81.8  &  26.27$\pm$3.0\\ \hline \\[-2ex]

\multirow{2}{*}{\textbf{B}} &  \multirow{2}{*}{X}   & \multirow{2}{*}{$(-8, 1, -45^o)$} & \multirow{2}{*}{$(8, 1, 0^o)$} &  GP-Frontier  & \textbf{21.12$\pm$0.6}  &  20.8$\pm$0.52  & \textbf{51.70$\pm$8.9} &  \textbf{10.03$\pm$12.36 } &  \textbf{13.8$\pm$0.6}\\ \\[-2ex]
 &     &   &  &  AG   & 22.85$\pm$0.8  &  \textbf{20.5$\pm$0.26}  & 66.36$\pm$15.2 &  25.68$\pm$26.53  &  14.8$\pm$0.9\\ \hline \\[-2ex]

\multirow{2}{*}{\textbf{B}} & \multirow{2}{*}{SU}   & \multirow{2}{*}{$(4, 4, 90^o)$} & \multirow{2}{*}{$(-2, -8, 0^o)$ }&  GP-Frontier  & $\mathbf{40.1\pm0.6}$  & \textbf{ 39.50$\pm$0.6}  & \textbf{46.20$\pm$6.1} &  \textbf{19.5$\pm$13.6}  & \textbf{ 28.2$\pm$1.6}\\ \\[-2ex]
  &    &   &  &  AG   & Fail  &  Fail  & Fail &  Fail  &  Fail\\ \hline \\[-2ex]

\multirow{2}{*}{\textbf{B}} & \multirow{2}{*}{CU}   &\multirow{2}{*}{ $(-8, 8, -45^o)$ }& \multirow{2}{*}{$(8, -8, 0^o)$} &  GP-Frontier  & \textbf{43.1$\pm$0.7}  &  \textbf{40.2$\pm$0.45}  & \textbf{48.1$\pm$6.9} &  \textbf{21.17$\pm$15.2}  & \textbf{ 27.9$\pm$1.1}\\ \\[-2ex]
 &    &  &   &  AG   & Fail  &  Fail  & Fail &  Fail  &  Fail\\ \hline \\[-2ex]

\multirow{2}{*}{\textbf{B}} &  \multirow{2}{*}{GU}   & \multirow{2}{*}{$(-5, -8, 45^o)$} & \multirow{2}{*}{$(4, 4, -90^o)$} &  GP-Frontier  & \textbf{30.9$\pm$0.5 } & \textbf{ 30.6$\pm$0.3}  & \textbf{41.3$\pm$5.6} &  \textbf{19.08$\pm$8.4}  &  \textbf{20.9$\pm$0.8}\\ \\[-2ex]
 &     &   &   &  AG   & Fail(110)     &  Fail  & Fail &  Fail  &  Fail\\ \hline  \\[-2ex]

\multirow{2}{*}{\textbf{Cafeteria}} & \multirow{2}{*}{-}   & \multirow{2}{*}{$(0,0,0^o)$} & \multirow{2}{*}{$(16, -4, 1^o)$} &  GP-Frontier   &$\mathbf{55.9\pm13.8}$    & $\mathbf{20.5\pm2.1}$  & $\mathbf{4.7\pm2.9}$ & $\mathbf{0.3\pm0.05}$  & $\mathbf{36.4\pm10.3}$ \\   \\[-2ex]
&     & & &  AG   & $83.9\pm17.9$     &  $22.6\pm4.7$ & $38.4\pm11.8$ &  $7.1\pm8.6$  & $98.9\pm24.3$ \\ \hline \\[-2ex]
\vspace{-2pt}
\end{tabular} \vspace{-8pt}
  \caption{Local Navigation Performance, improved metrics over the baseline is highlighted in bold for each experiment.\vspace{-2pt}}
  \label{tab_performance}
\end{table*}

\subsection{Simulation Results} \label{sec_sim_results}
Overall, the GP-Frontier and AG techniques were able to find a collision-free path for both the MD and X experiments, see Fig.~\ref{fig_env}. However, in the case of the SU, CU, and GU experiments, the AG approach failed to reach the goal (considering 10 trials).
Table \ref{tab_performance} shows that the average GP-Frontier and AG traveled distances for experiments MD and X are almost equal. Nevertheless, the GP-Frontier outperforms the AG in terms of total time, accumulated jerk, curvature change, and risk measure. 
The GP-Frontier generates smoother paths and decreases the risk value since the robot follows the center of the open space. However, these advantages (i.g. smoothness and distance from obstacles) may lead to slightly longer paths. For example, in experiment X, the AG's average traveled distance (20.5 m) is lower than that of the GP-Frontier (20.8 m), see Fig.~\ref{fig_exp_x} and Table~\ref{tab_performance}.
On the other hand, most of the paths generated by the AG methods include a point where the robot oscillates (either left and right or forward and backward) until it eventually selects a sub-goal, see the highlighted blue circle in Fig. \ref{fig_exp_md} and its corresponding blue square in Fig. \ref{fig_exp_md_w}. Fig.~\ref{fig_exp_md_time} shows the running time performance of VSGP model during experiment MD. The training time (green dots) required to train the VSGP model on the dataset $\mathcal{D}$ is below \num{20} milli-seconds for almost all observations, while the prediction time (blue dots) needed to estimate the probability of occupancy over the surface and its associated variance is around \num{60} milli-seconds for all observations.

The AG method encountered a local minimum in the remaining experiments: SU, CU, and GU, because it only takes into account the distance metric when selecting a sub-goal; it selects the gap that is nearest to the goal. 
In Exp. SU, at the initial position (indicated by a red circle in Fig. \ref{fig_exp_env2_b}), the robot can observe two gaps (located in the upper left and upper right corners of U1). The AG method chooses the upper left gap since it is the closest to the goal. However, as the robot departs from U1 towards the upper left gap, the lower edge of U1 goes out of the field of view, leading to the emergence of a new gap (lower gap) inside U1 that is now closest to the goal. The AG method subsequently selects the lower gap as the current sub-goal. Once the lower edge of U1 reappears, the AG method switches back to selecting the upper left gap as the sub-goal. This pattern of actions repeats continuously, as seen in Figs. \ref{fig_exp_env2_b} and \ref{fig_exp_env2_e}.
The situation is different for the GP-Frontier method because it selects a sub-goal based on both the distance and direction of the GP-Frontiers (gaps). So, Even when the new lower gap inside U1 appears, it will have a high cost in terms of direction with respect to the robot's heading. As a result, GP-Frontier will keep the upper left gap as the current sub-goal.
In  Exp. CU, the AG method encountered a similar problem as in  Exp. SU, getting stuck in a local minimum where the robot oscillates left and right inside U2, see Fig.~\ref{fig_exp_env2_c}.

In Exp. GU, the situation is different from that in Exp. SU and CU, as the AG gets stuck outside the U-shaped obstacles, however, the AG is stuck because of the same reason. When the robot reaches the lower right corner of U1, it encounters three gaps (with directions: up, down, and left (inside U3)). Since the upper and left gaps have approximately the same distance to the goal (inside U1), the AG method selects the gap nearest to the goal (assume the upper gap). But, as the robot moves towards the selected gap (following a curve around the lower right corner of U1), the other gap (left gap) will become the nearest to the goal. Thus, the AG method switched to it once again. This same sequence is repeated continuously, as depicted in Fig.~\ref{fig_exp_env2_d}. This behavior can be seen in the supplementary video. 

We believe that the AG method encountered local minima in environment \textbf{B} due to the challenging navigation through U-shaped rooms and wide walls, resulting in two gaps that are equidistant from the final goal. In contrast, environment \textbf{A} is less challenging because the obstacles' size is smaller. Our approach, unlike the AG method, takes into account both the direction and distance of the GP-Frontiers (gaps) in the cost function $C$, which minimizes the probability of getting trapped in a local minimum.

\begin{figure} \vspace{-10pt}
    \centering
  \subfloat[\label{fig_indoor_a}]{%
      \includegraphics[width=0.49\linewidth,height=1.1in]{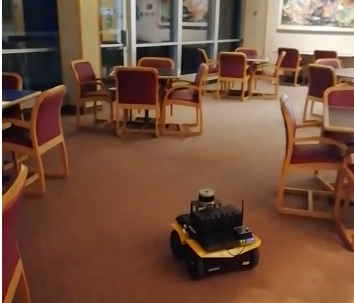} }
   \subfloat[\label{fig_indoor_b}]{%
      \includegraphics[width=0.49\linewidth,height=1.1in]{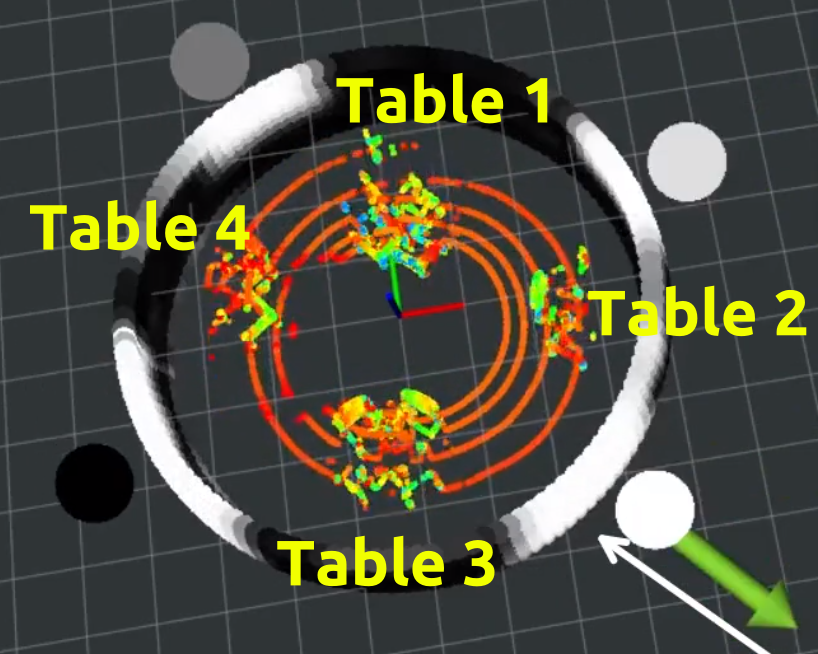} } \vspace{-10pt}
  \caption{\small Real robot evaluation in an indoor environment. (a) University cafeteria. (b) shows raw pointcloud, variance surface (grey-coded surface), detected GP-Frontiers which are visualized as grey-colored spheres; whiter spheres have less cost. The green arrow indicates the direction of the goal. \vspace{-20pt}}
  \label{fig_indoor} 
\end{figure}

\vspace{-5pt}
\subsection{Hardware Demonstration} \label{sec_sim_results}
\begin{figure} \vspace{-10pt}
    \centering
  \subfloat[\label{fig_outdoor_a}]{%
      \includegraphics[width=0.49\linewidth,height=1.1in]{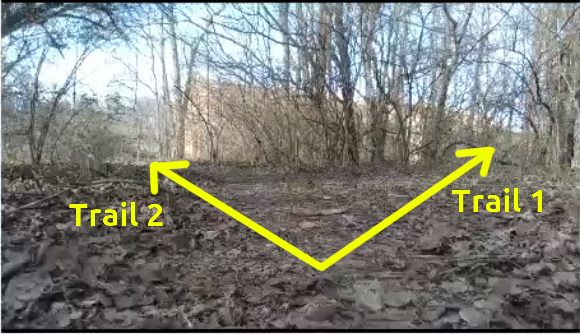} }
   \subfloat[\label{fig_outdoor_b}]{%
      \includegraphics[width=0.49\linewidth,height=1.1in]{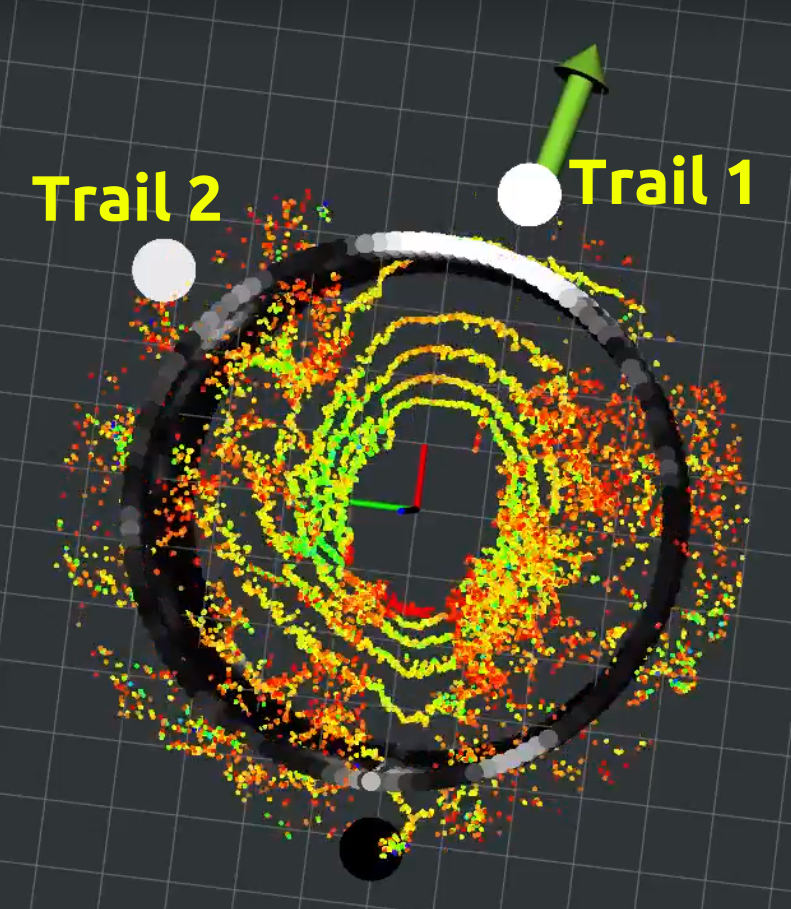} } \vspace{-10pt}
  \caption{\small Real robot evaluation in an outdoor environment (a) trail inside a forest. (b) shows raw pointcloud, variance surface (grey-coded surface), detected GP-Frontiers which are visualized as grey-colored spheres; whiter spheres have less cost. \vspace{-20pt}}
  \label{fig_outdoor} 
\end{figure}

A Jackal mobile robot, equipped with a VLP-16 LiDAR 
is used to validate our GP-Frontier approach. Our algorithm runs in real-time with a frequency of 10 Hz. 
Our GP-Frontier method is validated in both indoor and outdoor environments. The university cafeteria is chosen to represent a cluttered indoor environment similar to environment \textbf{A}, as shown in Fig.~\ref{fig_indoor}. In addition, the GP-Frontier is demonstrated in a harsh outdoor environment by testing it on a real forest trail, see Fig.~\ref{fig_outdoor}. 
Compared to other local gap-based navigation approaches, GP-Frontiers relies on the variance surface of the VSGP representation, which is more resilient to noisy measurements. This property is particularly valuable in noisy environments, as demonstrated in the noisy pointcloud generated in the forest, see Fig.~\ref{fig_outdoor_b}. 
By smoothing out the raw pointcloud observation, the variance surface enables better detection of the GP-Frontier (gaps) around the robot. The surface also exhibits a smoothing property for indoor environments, as evidenced by Fig.~\ref{fig_indoor}, where it smooths out the occupied and free spaces.
Specifically, the variance surface represents each table and its chairs as a single "big" obstacle, black regions on the surface, instead of being a sparse set of points as seen in the raw pointcloud data.

The performance table (Table~\ref{tab_performance}) demonstrates that our proposed approach outperforms the AG method in all performance metrics. The accumulated jerk $\mathrm{\mathbf{J}}_{\mathrm{\textbf{acc}}}$ and the trajectory curvature $\mathrm{\textbf{C}}_{\text {\textbf{chg} }}$ values in the real-world experiments are lower than those in the simulation experiment due to the maximum linear velocity being limited to \num{0.5} \si{\metre/\second} during the hardware demonstration.
Robot navigation in these environments with real-time computed GP-Frontiers can be watched in the supplementary video.

\vspace{-5pt}
\section{Conclusion} \label{conclusion}
\vspace{-3pt}
We present the GP-Frontier and its navigation approach to navigate the robot safely towards a goal without the need of any global map or path planner. 
The proposed approach is built on the uncertainty assessment of the VSGP occupancy model of the robot surrounding. The VSGP model provides a high-level representation of the environment, which is a more efficient representation to detect the open (safe) gaps around the robot and is more robust again the noisy measurement. The intensive evaluation shows that our approach has salient advantages over the most recent baseline method.

\vspace{-5pt}
\section*{ACKNOWLEDGMENT}
\vspace{-5pt}
This work was supported by the National Science Foundation with grant numbers 2006886 and 2047169.
We thank Hassan Jardali for his help during the field experiments. 

\bibliographystyle{unsrt}
\bibliography{references,ref2,ref3,ref-iros}

\begin{thebibliography}{10}

\bibitem{yamauchi1997frontier}
Brian Yamauchi.
\newblock A frontier-based approach for autonomous exploration.
\newblock In {\em Proceedings 1997 IEEE International Symposium on
  Computational Intelligence in Robotics and Automation CIRA'97.'Towards New
  Computational Principles for Robotics and Automation'}, pages 146--151. IEEE,
  1997.

\bibitem{yamauchi1998frontier}
Brian Yamauchi.
\newblock Frontier-based exploration using multiple robots.
\newblock In {\em Proceedings of the second international conference on
  Autonomous agents}, pages 47--53, 1998.

\bibitem{holz2010evaluating}
Dirk Holz, Nicola Basilico, Francesco Amigoni, and Sven Behnke.
\newblock Evaluating the efficiency of frontier-based exploration strategies.
\newblock In {\em ISR 2010 (41st International Symposium on Robotics) and
  ROBOTIK 2010 (6th German Conference on Robotics)}, pages 1--8. VDE, 2010.

\bibitem{shen2012stochastic}
Shaojie Shen, Nathan Michael, and Vijay Kumar.
\newblock Stochastic differential equation-based exploration algorithm for
  autonomous indoor 3d exploration with a micro-aerial vehicle.
\newblock {\em The International Journal of Robotics Research},
  31(12):1431--1444, 2012.

\bibitem{butzke20153}
Jonathan Butzke, Andrew Dornbush, and Maxim Likhachev.
\newblock 3-d exploration with an air-ground robotic system.
\newblock In {\em 2015 IEEE/RSJ International Conference on Intelligent Robots
  and Systems (IROS)}, pages 3241--3248. IEEE, 2015.

\bibitem{wang2018collaborative}
Luqi Wang, Daqian Cheng, Fei Gao, Fengyu Cai, Jixin Guo, Mengxiang Lin, and
  Shaojie Shen.
\newblock A collaborative aerial-ground robotic system for fast exploration.
\newblock In {\em International Symposium on Experimental Robotics}, pages
  59--71. Springer, 2018.

\bibitem{burgard2000collaborative}
Wolfram Burgard, Mark Moors, Dieter Fox, Reid Simmons, and Sebastian Thrun.
\newblock Collaborative multi-robot exploration.
\newblock In {\em Proceedings 2000 ICRA. Millennium Conference. IEEE
  International Conference on Robotics and Automation. Symposia Proceedings
  (Cat. No. 00CH37065)}, volume~1, pages 476--481. IEEE, 2000.

\bibitem{cieslewski2017rapid}
Titus Cieslewski, Elia Kaufmann, and Davide Scaramuzza.
\newblock Rapid exploration with multi-rotors: A frontier selection method for
  high speed flight.
\newblock In {\em 2017 IEEE/RSJ International Conference on Intelligent Robots
  and Systems (IROS)}, pages 2135--2142. IEEE, 2017.

\bibitem{1266644}
Javier Minguez and L.~Montano.
\newblock Nearness diagram (nd) navigation: collision avoidance in troublesome
  scenarios.
\newblock {\em IEEE Transactions on Robotics and Automation}, 20(1):45--59,
  2004.

\bibitem{mujahed2018admissible}
Muhannad Mujahed, Dirk Fischer, and B{\"a}rbel Mertsching.
\newblock Admissible gap navigation: A new collision avoidance approach.
\newblock {\em Robotics and autonomous systems}, 103:93--110, 2018.

\bibitem{sezer2012novel}
Volkan Sezer and Metin Gokasan.
\newblock A novel obstacle avoidance algorithm:“follow the gap method”.
\newblock {\em Robotics and Autonomous Systems}, 60(9):1123--1134, 2012.

\bibitem{ye2009sub}
Chen Ye and Phil Webb.
\newblock A sub goal seeking approach for reactive navigation in complex
  unknown environments.
\newblock {\em Robotics and Autonomous Systems}, 57(9):877--888, 2009.

\bibitem{rasmussen2002infinite}
Carl Rasmussen and Zoubin Ghahramani.
\newblock {Infinite Mixtures of Gaussian Process Experts}.
\newblock In {\em Advances in Neural Information Processing Systems}, 2002.

\bibitem{snelson2006sparse}
Edward Snelson and Zoubin Ghahramani.
\newblock Sparse gaussian processes using pseudo-inputs.
\newblock {\em Advances in neural information processing systems}, 18:1257,
  2006.

\bibitem{titsias2009variational}
Michalis~K Titsias.
\newblock Variational model selection for sparse gaussian process regression.
\newblock {\em Report, University of Manchester, UK}, 2009.

\bibitem{hoang2015unifying}
Trong~Nghia Hoang, Quang~Minh Hoang, and Bryan Kian~Hsiang Low.
\newblock {A unifying framework of anytime sparse Gaussian process regression
  models with stochastic variational inference for big data}.
\newblock In {\em International Conference on Machine Learning}, pages
  569--578. PMLR, 2015.

\bibitem{bauer2016understanding}
Matthias Bauer, Mark van~der Wilk, and Carl~Edward Rasmussen.
\newblock {Understanding probabilistic sparse Gaussian process approximations}.
\newblock In {\em Advances in neural information processing systems}, pages
  1533--1541, 2016.

\bibitem{ma2017informative}
Kai-Chieh Ma, Lantao Liu, and Gaurav~S Sukhatme.
\newblock Informative planning and online learning with sparse gaussian
  processes.
\newblock In {\em 2017 IEEE International Conference on Robotics and Automation
  (ICRA)}, pages 4292--4298. IEEE, 2017.

\bibitem{kim2012building}
Soohwan Kim and Jonghyuk Kim.
\newblock Building occupancy maps with a mixture of gaussian processes.
\newblock In {\em 2012 IEEE International Conference on Robotics and
  Automation}, pages 4756--4761. IEEE, 2012.

\bibitem{luo2018adaptive}
Wenhao Luo and Katia Sycara.
\newblock Adaptive sampling and online learning in multi-robot sensor coverage
  with mixture of gaussian processes.
\newblock In {\em 2018 IEEE International Conference on Robotics and Automation
  (ICRA)}, pages 6359--6364. IEEE, 2018.

\bibitem{GPOM}
Simon~T O’Callaghan and Fabio~T Ramos.
\newblock Gaussian process occupancy maps.
\newblock {\em The International Journal of Robotics Research}, 31(1):42--62,
  2012.

\bibitem{jadidi2018gaussian}
Maani~Ghaffari Jadidi, Jaime~Valls Miro, and Gamini Dissanayake.
\newblock Gaussian processes autonomous mapping and exploration for
  range-sensing mobile robots.
\newblock {\em Autonomous Robots}, 42(2):273--290, 2018.

\bibitem{realtime-expo}
Wennie Tabib, Kshitij Goel, John Yao, Mosam Dabhi, Curtis Boirum, and Nathan
  Michael.
\newblock Real-time information-theoretic exploration with gaussian mixture
  model maps.
\newblock In {\em Robotics: Science and Systems}, 2019.

\bibitem{ali2023light}
Mahmoud Ali and Lantao Liu.
\newblock Light-weight pointcloud representation with sparse gaussian process.
\newblock {\em arXiv preprint arXiv:2301.11251}, 2023.

\bibitem{yan2020mapless}
Kai Yan and Baoli Ma.
\newblock Mapless navigation based on 2d lidar in complex unknown environments.
\newblock {\em Sensors}, 20(20):5802, 2020.

\bibitem{matthews2017gpflow}
Alexander G de~G Matthews, Mark Van Der~Wilk, Tom Nickson, Keisuke Fujii,
  Alexis Boukouvalas, Pablo Le{\'o}n-Villagr{\'a}, Zoubin Ghahramani, and James
  Hensman.
\newblock Gpflow: A gaussian process library using tensorflow.
\newblock {\em J. Mach. Learn. Res.}, 18(40):1--6, 2017.

\end{thebibliography}

\end{document}